\documentclass[publish,paper]{jaciiiarticle}
\usepackage[dvips]{graphicx} 
\usepackage{epsfig}  
\usepackage[T1]{fontenc}
\usepackage{amsfonts,amsmath,amssymb,bm} 
\usepackage{algorithm}
\usepackage{algorithmic}
\usepackage{graphicx}
\newcommand{\Frechet}{Fr\'echet}

\begin{document}

\pagestyle{jaciiistyle}

\title{Data-driven Analysis for Understanding Team Sports Behaviors} 
\author{Keisuke Fujii}
\address{Nagoya University, Furocho 1, Nagoya, Aichi, 464-8603, JAPAN\\
         E-mail: fujii@i.nagoya-u.ac.jp}
\markboth{Keisuke Fujii}{Machine Learning based Analysis for Team Sports Behaviors}
\maketitle

\begin{abstract}
Understanding the principles of real-world biological multi-agent behaviors is a current challenge in various scientific and engineering fields. 
The rules regarding the real-world biological multi-agent behaviors such as team sports are often largely unknown due to their inherently higher-order interactions, cognition, and body dynamics.
Estimation of the rules from data, i.e., data-driven approaches such as machine learning, provides an effective way for the analysis of such behaviors. 
Although most data-driven models have non-linear structures and high prediction performances, it is sometimes hard to interpret them.
This survey focuses on data-driven analysis for quantitative understanding of invasion team sports behaviors such as basketball and football, and introduces two main approaches for understanding such multi-agent behaviors: (1) extracting easily interpretable features or rules from data and (2) generating and controlling behaviors in visually-understandable ways.
The first approach involves the visualization of learned representations and the extraction of mathematical structures behind the behaviors.
The second approach can be used to test hypotheses by simulating and controlling future and counterfactual behaviors. 
Lastly, the potential practical applications of extracted rules, features, and generated behaviors are discussed. 
These approaches can contribute to a better understanding of multi-agent behaviors in the real world.
\end{abstract}

\begin{keywords}
Human behavior, Machine learning, Dynamical systems, Sports, Interpretability
\end{keywords}

\section{Introduction}
The development of measurement technologies has made possible the measurement and analysis of the movements of various organisms. 
For example, they have enabled an understanding of the behaviors of wild animals and athletes from data.
Specifically, recent advances in sports-related measurement technologies have been reviewed by many researchers such as in \cite{Rahmad18, Kamble19, Rico20}. 
Based on the advances, it is now possible to obtain a better understanding of the principles of real-world biological multi-agent behaviors, which is a fundamental problem in various scientific and engineering fields. 
The rules underlying real-world biological multi-agent behaviors are often largely unknown because the elements are not physically connected. 
Mathematical models based on simple rules are used to directly understand the multi-agent movements. 
For example, models based on social forces are widely applied, in which a force is assumed to be acting among individuals \cite{Helbing95}.
In a limited number of situations, these models are also applied to more complicated behaviors such as sports \cite{Yokoyama18, Spearman17, Alguacil20}. However, modeling the general multi-agent behaviors of living organisms in the real world (e.g., team sports) can be mathematically difficult due to their inherently higher-order social interactions, cognition, and body dynamics \cite{Fujii16}. 
Therefore, to obtain a better understanding of these behaviors, a data-driven and model-free (or equation-free) approach \cite{Fujii18, Fujii19c} is needed.

Data-driven modeling is a powerful approach such as for extracting information and making a prediction using complex real-world data.
For example, learning models with complex nonlinear structures such as neural networks, are actively studied in the field of machine learning.
Although these nonlinear models are often effective in terms of obtaining higher expressiveness and predictive performance, they are sometimes difficult to interpret. 
Hence, this study aims to bridge the gap between rule-based (or traditional sports sciences) and data-driven approaches, for which there is a trade-off between interpretability and expressiveness (or predictability). 
The next questions must be: what kind of nonlinear data-driven model will enable a better quantitative understanding? 
In a discussion of this issue regarding the relationship between cognitive science and deep neural network models \cite{Cichy19}, the authors mentioned that such models would have value if they could predict and explain phenomena, which could serve as a starting point for the establishment of new theories.
In the case of complex multi-agent behaviors, existing rule-based models are too simple.
To obtain a better understanding, indirect techniques using nonlinear data-driven models are required: e.g., (i) extracting the mathematical structure behind the motions, (ii) visualizing the learned representations, and (iii) modeling the components and generating plausible motions. 
If this requirement can be satisfied, even the results are based on a nonlinear data-driven model, it will be possible to contribute to the understanding of complex multi-agent behaviors.

In this paper, data-driven analyses for team sports behaviors are introduced, especially in invasion sports such as basketball and football, which show complex interactive behaviors.  
A range of related surveys or dissertations have addressed the spatio-temporal aspects of this issue \cite{Gudmundsson17,Horton18,Sha18} with a focus on football \cite{Tuyls20}, and have discussed prediction approaches via machine learning \cite{Beal19, Bunker19, Keshtkar19} including match outcome prediction, tactical decision making, player investments, fantasy sports, and injury prediction.
The contribution of this paper is to review data-driven analyses that interpret team sports behaviors (e.g., based on the trajectory and action data of the players and ball as defined in Section \ref{sec:preliminary}), rather than simply performing clustering, classification, and prediction via black-box learning-based models.
After the preliminary explanation of terms in Section \ref{sec:preliminary}, examples of data-driven approaches to extract features and rules are introduced in Section \ref{sec:extract}, including the visualization of learned representations and extraction of mathematical structures underlying the behaviors. 
In Section \ref{sec:simulate}, an approach for testing hypotheses by simulating and controlling plausible future behaviors by generating future and counterfactual behaviors is introduced. Lastly, in Section \ref{sec:practical}, the potential for the practical application of these estimated rules, features, and generated predictions is discussed. 

\begin{figure}[h]
\centering
\includegraphics[width=0.6\columnwidth]{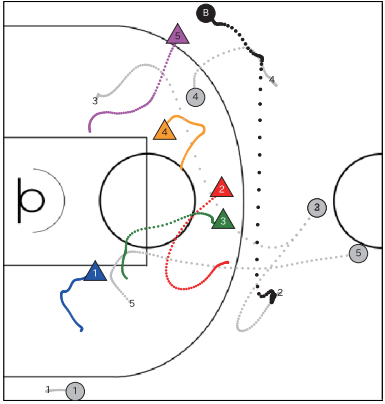}
\caption{An example of multi-agent trajectory data in basketball (illustration from \cite{Fujii20policy}). The colored triangles, gray circles, and the black circle represent the defenders, attackers, and ball, respectively.}
\label{fig:example}
\end{figure}

\begin{figure}[h]
\centering
\includegraphics[width=0.7\columnwidth]{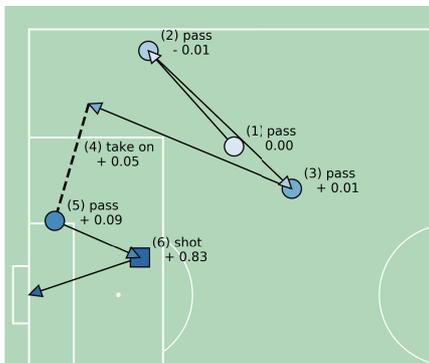}
\caption{An example of a player action sequence with a ball in soccer (illustration from \cite{Decroos19}).}
\label{fig:example_soccer}
\end{figure}


\section{Preliminary}
\label{sec:preliminary}
The term \emph{agent} is used to denote a dynamic object of interest such as a player or the ball in team sports.
\emph{A single-agent trajectory} $P$ of length $m$ is a sequence of $m$ features $P = (p_1,p_2,\dots,p_{m})$, where $p_i \in \mathbb{R}^d$ is a features with $d$ dimensions.
For example, as a feature, the $d$-dimensional coordinate is a simple case.
\emph{Multi-agent trajectories} $P_K$ (e.g., a team, both teams, or with the ball, such as in Figure \ref{fig:example}) with $K$ agents comprise a sequence of $m$ features $P_K = (p_{K,1},p_{K,2},\dots,p_{K,m})$, where $p_{K,i} = [p_{i,1},p_{i,2},\dots p_{i,K}] \in \mathbb{R}^{K \times d}$.
A sequence of relations in a multi-agent system $\bm{R}_K$ is defined as $\bm{R}_K = (R_1,R_2,\dots,R_{m})$, where $R_i \in \mathbb{R}^{K \times K}$.
$R_i$'s component $R_{i,k,l}$ represents the relation between agents $k$ and $l$ at each $i$, as computed by $R_{i,k,l} = h(p_{i,k},p_{i,l}) $ (e.g., $h$ is a distance function \cite{Fujii17,Fujii18} or a Gaussian kernel \cite{Fujii19b,Fujii20}).
In the following, \emph{actions} indicate discrete behaviors such as dribble, pass, and shot, as shown in Figure \ref{fig:example_soccer}. 
The objective of this paper is to present a method for obtaining a better understanding of team sports \emph{behaviors}, including continuous trajectories and discrete actions.

\section{Extracting features and rules from data}
\label{sec:extract}
This learning-based approach is used to extract features and rules despite the availability of little prior knowledge.
In this section, conventional rule-based approaches are firstly introduced, followed by unsupervised and supervised learning approaches, with particular regard to their interpretation.
The unsupervised and supervised approaches comprise two of the three main categories of machine learning (the third is reinforcement learning, which is introduced in Section \ref{sec:simulate}).

\subsection{Conventional rule-based approaches}
\label{ssec:handcrafted}
In conventional methods without learning-based approaches, researchers in various fields have evaluated the characteristics of multi-agent behaviors based on their experience and established theories. 
For example, based on hypotheses, researchers have calculated the distances and relative phases of two athletes (e.g., \cite{Bourbousson10, Travassos12, Fujii16}), the speeds of movements (e.g., \cite{Sampaio15}), the frequencies and angles of actions (e.g., shots \cite{Goldsberry12} and passes \cite{Correia11,Vilar13,Fujii20cognition}), and their representative values (e.g., average and maximum values). 
Measurement systems with greater spatiotemporal resolution (e.g., motion capture systems and force platforms) can analyze skillful maneuvers \cite{Fujii14,Fujii15} in terms of their cognition \cite{Brault12, Fujii14b}, force \cite{Fujii15b}, and torque \cite{Fujii15c}.
After obtaining representative values, specific hypotheses have been tested (e.g., \cite{Fujii16,Power18}) sometimes by statistical analysis.
For example, in order to quantify the flexible teamwork of basketball defense (i.e., 5-vs-5), evaluation of the defensive cooperation against team attacks called screen-plays, which block the movements of a defender, was performed \cite{Fujii16}. The results showed that the defender flexibly changes the frequency of the four roles (i.e., switching, overlapping, ignoring, global-help) according to the level of the emergency.
This traditional quantitative approach remains powerful, is applicable to small datasets, and is the easiest to interpret in a range of fields (e.g., a particular sport) because it allows for the direct test of the hypothesis.

Representative values have also been computed using more mathematically sophisticated approaches.
Pioneering work was conducted in which each player's area of control in actual soccer games was evaluated as a Voronoi diagram \cite{Taki00}. 
Other studies, for example, have analyzed the connection of passes based on network theory \cite{Yamamoto11}, the self-similarity hidden in a time series of the front position of the team \cite{Kijima14}, and the breaking of spatiotemporal symmetry using group theory in a 3-vs-1 ball possession task \cite{Yokoyama11}. 
In a recent work \cite{Spearman18}, a probabilistic physics-based model was developed to quantify off-ball scoring opportunities. 
However, in order to represent cooperative/competitive interactions in a more detailed or practical manner, more flexible modeling would be needed.
A wide variety of data-driven methods such as using machine learning have basically been developed to achieve clear objectives such as automatic feature extraction, classification, and regression.
In the following subsection, unsupervised and supervised learning, which are used in the field of machine learning \cite{Bishop06} are introduced, and examples of researches using player position data in team sports, which can be readily interpreted, are presented. 

\subsection{Unsupervised learning}
\label{ssec:unsupervised}
Unsupervised learning involves a type of machine learning algorithm that acquires insight by inferring a function for describing hidden structures from unlabeled data.
This is a powerful approach for knowledge discovery from data without the benefit of clear hypotheses. 
Typical unsupervised methods include dimensionality reduction and clustering. 
Dimensionality reduction is the transformation of high-dimensional data into a meaningful representation of lower dimensionality.
For example, principal component analysis (PCA), t-distributed stochastic neighbor embedding (t-SNE) \cite{Maaten08} regarding shot type \cite{Marty18}, non-negative matrix factorization (NMF) \cite{Miller14} or tensor decomposition \cite{Papalexakis18} regarding the shot type, and topic modeling \cite{Wang15,Miller17} of the trajectory (i.e., one of the natural language processing algorithms), have been used to summarize diverse interactive sports behaviors into lower-dimensional representations. 
However, some of these methods have assumed independence of sampling. 
That is, the extracted information does not reflect dynamical properties. 
Therefore, an extraction method identifying the coordinative structures based on dynamical properties from data is needed. 

A number of approaches are used to reduce the number of dimensions while considering the time-series structures.
For example, image-based approaches transform trajectory data into images using neural networks (e.g., \cite{Wang16,Nistala18}), including the self-organizing map (e.g., \cite{Grunz12,Kempe15}).
Another approach for extracting physically-interpretable dynamical information is a method called \emph{dynamic mode decomposition} (DMD) \cite{Rowley09, Schmid10}.
It can decompose data into a small number of time dynamics (i.e., frequency and growth rate) and their coefficients (i.e., extraction of dynamic properties).
DMD is based on the spectral theory of the Koopman operator \cite{Koopman31,Mezic05}. Theoretically, to compute DMD, the data must be rich enough to approximate the eigenfunctions of the Koopman operator. 
However, in basic DMD algorithms that naively use the obtained data, the above assumption is not satisfied e.g., when the data dimension is too small to approximate the eigenfunctions. 
Thus, there are several algorithmic variants of DMDs to overcome this problem such as a formulation in reproducing kernel Hilbert spaces (RKHSs) \cite{Kawahara16}, in a multitask framework \cite{Fujii19a}, and using a neural network \cite{Takeishi17c}. 
Researchers have applied the DMD in RKHSs to multi-agent relation sequences (see Section \ref{sec:preliminary} in team sports \cite{Fujii17,Fujii18} and utilized the structure of an adjacency matrix series $\bm{R}_K$ (see Section \ref{sec:preliminary}) via tensor-train decomposition \cite{Fujii20}.
This approach has the advantage of enabling (i) the extraction of the mathematical structure and (ii) visualization of the learned expressions for the above purposes of data-driven methods.


Clustering involves grouping a set of objects such that objects in the same group (called a cluster) are more similar to each other than to those in other groups (clusters).
There are many clustering algorithms based on various cluster models, e.g., hierarchical clustering (based on the connectivity or similarity between two trajectories), centroid-based clustering (such as k-means), and distribution-based clustering (such as Gaussian mixture models).
For team sports data, researchers have used hierarchical clustering \cite{Hobbs18,Hobbs19} based on similarity \cite{Decroos18,Sha16,Kanda20} and distribution-based clustering using a Gaussian mixture model \cite{Pervse09}.
However, again, problems can occur when using time-series data (for example, it is difficult to naively compute a similarity when the data do not have fixed time lengths). 
In that case, one approach is to specifically design the similarity of time series to enable the application of the conventional clustering method to static data. 

Hierarchical clustering requires appropriate distance measures.
Among the several distance measures available for trajectories, the \Frechet{} distance \cite{Alt95} and dynamic time warping (DTW) \cite{Berndt94} have been frequently used (e.g., in basketball \cite{Decroos18,Sha16} soccer \cite{Decroos18} games).
However, these simple approaches have high computational costs and are difficult to apply to large-scale sports data. 
Therefore, researchers have developed a scalable method for computing \Frechet{} distance by quickly performing a search on a tree data structure called \emph{trie} \cite{Kanda20}.
Recently developed neural network approaches can also compute the similarities of a single-agent trajectory in scalable ways \cite{Yao17,Li18}, but these have not been applied to team sports multi-agent trajectory datasets. 

Another problem is the computation of the distance or similarity between multi-agent trajectories.
A simple method for comparing agent-to-agent trajectories encounters permutation problems among the players \cite{Sha16}.
One rule-based approach permutes the players nearest to the ball used such as in \cite{Fujii18,Fujii20}. 
A data-driven permutation method such as a linear assignment, known as the Hungarian algorithm \cite{Papadimitriou82}, has also been used for role assignment problems in Basketball  \cite{Lucey13representing,Lucey14how,Sha16,Sha17} and soccer \cite{Le17,Yeh19} (e.g., guard, forward, and center in basketball).

Another approach to deal with the permutation problem is calculating the similarity of multivariate nonlinear dynamical systems using DMD \cite{Fujii17, Fujii18}.
Since DMD is a dimensionality reduction method like PCA, the extracted dynamical property is permutation-invariant. 
Moreover, this approach uses a kernel that reflects the dynamics via the extraction of dynamical properties.
A kernel called the Koopman spectral kernel can be regarded as a similarity between multivariate nonlinear dynamical systems, which permits the use of some clustering methods.
However, in general, since unsupervised learning methods do not use objective variables (labeled data), it is sometimes difficult to validate them quantitatively.
To evaluate them quantitatively, combining them with the following supervised learning methods may be effective.

\subsection{Supervised learning}
\label{ssec:supervised}
Supervised learning is a machine learning task of inferring a function from supervised or labeled training data.
When labeled data has discrete values such as the type of play, it is called classification, and when it has (relatively) continuous values such as position and score, it is referred to as regression.
Here, classification problems of team plays or regression problems for scoring probability are considered (other regression problems such as trajectory prediction are described later).
A simple approach is to input static features into classification or regression models. 
For example, score prediction in basketball
\cite{Cervone14,Chang14,Lucey14quality}, team identification in soccer \cite{Lucey13assessing}, screen-play classification,  \cite{Mcqueen14,Mcintyre16,Hojo18}, and prediction of who will obtain a basketball in rebounding situations \cite{Hojo19} using such as linear discriminant analysis (LDA), logistic regression, or support vector machine (SVM) with the hand-crafted static features described in Section \ref{ssec:handcrafted}.
In this process, the static features obtained from unsupervised learning (e.g., \cite{Bialkowski14}) can be input into classification or regression models. 

However, it is often necessary to reflect the time-series structure also when supervised learning is applied to complex multi-agent behaviors. 
A simple approach is to use the dynamic features obtained from unsupervised learning, as described in Section \ref{sec:extract}.  
For example, by the use of the above DMD and computation of similarity, defensive tactics (defending the area or players) and offensive tactics (with or without cooperation) \cite{Fujii20} can be classified. 
Another supervised learning method has also been used to classify and predict the scoring probability \cite{Fujii17, Fujii18}. The strength of supervised learning is that the results can be clearly evaluated.

More sophisticated approaches are end-to-end approaches, which use the same model to extract features and perform predictions (i.e., classification or regression).
For example, a neural network approach can be used to classify 
offensive plays \cite{Wang16}, team styles \cite{Mehrasa18}, and attack outcomes based on evaluating micro-actions \cite{Sicilia19}.
However, a neural network approach sometimes lacks interpretability. 
To obtain interpretable spatial representations, researchers have developed a number of approaches that provide both predictability and interpretability, such as using matrix \cite{Yue14} and tensor \cite{Zheng18,Park20} factor models, and Poisson point process \cite{Mortensen19}.
Other researchers have applied a supervised pattern mining method to rugby event data \cite{Bunker20}, which can also be applied to trajectory data after transforming the data into symbol sequences.

The combinations of the predictability and interpretability are related to practical applications to actual sports games because coaches and players need information such as why the score was obtained and what characteristics are observed in the subsequent plays. 
To explain and understand multi-agent behaviors more quantitatively or practically, it is necessary not only to improve prediction performance, but also to clarify their underlying principles (e.g., identify the mathematical structure and provide visualized representations that are interpretable).
Meanwhile, if the purpose of an analysis is close to its practical application, such as simulating and controlling behaviors as discussed in the next Section \ref{sec:simulate}, there may be no problems in using even black-box learning-based models.

\section{Simulating and controlling behaviors}
\label{sec:simulate}
This approach enables verification of researchers' hypothesis by modeling for future prediction or in situations that cannot be actually measured. 
In this section, conventional rule-based (or physics-based) approaches are introduced, followed by pattern-based (or data-driven) and planning-based approaches, based on the categorization of a human trajectory prediction survey \cite{Rudenko20}.
Pattern-based methods approximate an arbitrary dynamics function from training data to discover statistical behavioral patterns.
Planning-based methods explicitly address long-term movement goals of an agent and compute policies or path hypotheses that enable the agent to reach those goals (often formulated as reinforcement learning). 

\subsection{Conventional rule-based approaches} 
Traditionally, rule-based (or physics-based) methods enable researchers to determine and model the parameters of models (e.g., player position, speed, and interaction with other players). 
For example, the movements of players in a 3-vs-1 soccer possession task was simply modeled using three virtual social forces: spatial, avoiding, cooperative forces \cite{Yokoyama18}. 
In actual soccer games, pass probabilities \cite{Spearman17} and the future trajectories of players in several seconds \cite{Alguacil20} have been modeled using more complex rule-based approaches. 
These approaches have the advantage of providing an understanding of simulated and controlled behaviors because the users set all of the parameters. However, the adaptation of this approach to different problems (e.g., from soccer to basketball) requires additional and costly human labor. 

\subsection{Pattern-based approaches}
\label{ssec:pattern}
Pattern-based or data-driven approaches learn dynamics from data using less human knowledge to solve the above problem. 
In studies of team sports, there have been mainly two goals in applying these approaches: simulating multi-agent trajectories over several seconds and a more long-term team outcome. 
To predict long-term outcomes, if short-term behaviors are ignored, it is possible to simulate behaviors until the end of the possession (or attack).
Although this methodology mainly involves supervised learning, which overlaps with the content in Section \ref{ssec:supervised}, these methods are used to simulate and evaluate player behaviors, rather than extracting features and rules. 
In particular, researchers can use the reinforcement learning framework to evaluate either a player's action and state, or the team state to achieve the goals described in the following Section \ref{ssec:planning}.
In this subsection, modeling methods of multi-agent trajectories are then introduced.

\subsubsection{Simulating multi-agent trajectories}
The prediction of even just a few seconds of the multi-agent trajectories in team sports, e.g., basketball and soccer, is generally difficult.
That is why it is one of the benchmark problems in the field of machine learning \cite{Zheng16,Le17,Zhan19,Yeh19,Liu19naomi}.
Most methods have leveraged recurrent neural networks (RNN) \cite{Zheng16,Le17,Seidl18,Ivanovic18} including variational RNNs \cite{Zhan19,Yeh19}, although some have utilized generative adversarial networks (e.g., \cite{Chen18generating,Hsieh19}) and variational autoencoders \cite{Felsen18} without RNNs. 
Most of these methods were simply formulated as a trajectory prediction problem, whereas a few studies formulated it as an imitation learning problem (e.g., \cite{Le17,Fujii20policy}, which is one of reinforcement learning framework utilizing demonstration of experts (i.e., data).  

Most of these methods assume full observation to achieve long-term prediction in a centralized manner (e.g., \cite{Zhan19,Yeh19}).
In such a case, an important latent factor, e.g., whose information is utilized by each agent, is not interpretable.
Methods for learning attention-based observation of agents have been proposed for multi-agent in virtual environments and in real-world systems \cite{Hoshen17,Li20,Fujii20policy}.
Other approaches such as relational (e.g., \cite{Kipf18,Graber20}) 
or a physically-interepretable approaches \cite{Fujii19b,Fujii20} 
can learn interpretable representations of interactions.
Rigorously, decentralized modeling \cite{Fujii20policy} is needed to enable computation of each agent observation (or contribution).
Meanwhile, recent graph neural network approaches can learn permutation-equivariant features \cite{Kipf18,Yeh19,Sun19,Graber20}, which solve the permutation problem described in Section \ref{ssec:unsupervised}.

Another important approach is the tactical evaluation of a predicted trajectory.  
For example, trajectory prediction reflecting defensive evaluations in soccer \cite{Teranishi20} and trajectory computation optimizing defensive evaluations in basketball \cite{Sha18}.
Qualitatively, the evaluation of counterfactual prediction (i.e., predicting "what if" situations) can be used to validate the models \cite{Yeh19,Fujii20policy} based on the user's knowledge, whereas there is no ground truth in a counterfactual situation.

Although it is generally difficult to extract mathematical structures with such an approach that prioritizes predictive performance and performs a nonlinear transformation, there are methods that make them compatible such as in \cite{Fraccaro17} with applications other than sports.
Such methods can be useful for explicit modeling (e.g., observation model) of the nonlinear model when the phenomenon can be sufficiently explained or used as a starting point for various theories \cite{Cichy19} as mentioned in the Introduction. 
These approaches enable realistic and visually-understandable simulations (e.g., average athlete movements and the response to unobserved movements). 
Potential practical applications are presented in the following Section \ref{sec:practical}.


\subsection{Planning-based approaches}
\label{ssec:planning}
Planning-based methods explicitly address the long-term movement goals of agents and compute policies or path hypotheses that enable the agent to reach those goals. 
According to \cite{Rudenko20}, planning-based approaches are classified into two categories: inverse and forward planning methods.
Inverse planning methods estimate the action model or reward function from observed data using statistical learning techniques.
In other words, this approach utilizes a 
reinforcement learning framework in physical spaces (or in real-world data). 
Although it sometimes overlaps with supervised learning in Section \ref{ssec:supervised} and imitation learning in Section \ref{ssec:pattern}, the methods introduced here are used to evaluate actions and states of a player or a team to achieve their goals, rather than to extract features and rules or predict trajectories. 
Forward planning methods make an explicit assumption regarding the optimal criteria of an agent's movements, using a pre-defined reward function (e.g., a score in team sports). 
These two approaches are described in this subsection.

\subsubsection{Inverse approach using real-world data}
\label{sssec:realrl}

The inverse planning approach uses statistical learning techniques to estimate an action model or reward function from observed data.
Here this idea is extended to consider and value players' actions and the team's states.
For example, with respect to shooting, valuing player's actions by estimating the scoring and conceding probability (VAEP) \cite{Decroos19} and estimating a state-action value function (Q-function) using an RNN \cite{Liu18,Liu20}, which made interpretable using a linear model tree \cite{sun20}.
To evaluate the shooting action of players, researchers have investigated allocative efficiency in shot \cite{Sandholtz19}, the expected possession value  \cite{Cervone14,Cervone16multiresolution,Fernandez19}, and the value of the space \cite{Cervone16NBA,Fernandez18} by extending a Voronoi diagram \cite{Taki00}.
Regarding passing actions, similarly, researchers have used modeling and valuing of a pass \cite{Power17,Goes19,Bransen19}, pass-receiving \cite{Llana20}, the defender's pass-interception \cite{Robberechts19}.
In team plays, deep reinforcement learning to estimate the quality of the defensive actions was used in ball-screen defense in basketball \cite{wang18}. 
Another approach is the qualitative evaluation of counterfactual predictions as described above. 
For example, researchers have modeled the transition probabilities and shot policy tensors and have simulated seasons under alternative shot policies of interest \cite{Sandholtz18}.

\subsubsection{Forward approach in virtual spaces}
The forward planning approach involves the development of algorithms for the purpose of winning a competition involving humans or machines in virtual space (e.g., video games). 
To develop methods both in physical and virtual spaces, RoboCup (the Robot World Cup Initiative) involves attempts by robot teams to actually play a soccer game \cite{Kitano97}.
Research has been conducted on virtual games such as puzzles and shooters, and recently an open-source simulator for soccer games was published \cite{Kurach20}. 
Some researchers used a 3-vs-3 basketball simulator \cite{Tang18}, which is not currently open-source.
In these studies, using reinforcement learning, the performances are expected to defeat humans in some cases (such as mastering the game of Go \cite{Silver16}). 
It is also possible to learn similar behaviors from measurement data in sports games (e.g., using imitation learning frameworks as mentioned above). 
However, a few studies have combined inverse and forward planning-based frameworks.
For example, reinforcement learning could generate the optimal defensive team trajectory with the reward of preventing opponent scores after the imitation learning \cite{Sha16}.
An approach to bridge this gap is an important issue for future research.

\section{Practical applications and future directions}
\label{sec:practical}
There are a number of possible practical applications of extracted rules, features, and generated behaviors.
First, if play classification and score prediction become possible based on the extracted features and rules of multi-agent behaviors, the most directly useful application is the decrease in the workload of those who manually classified and evaluated plays by watching videos. 
However, it would be sometimes difficult to define specific plays that the user wishes to classify, whereas other plays can be easily defined (e.g., offensive and defensive tactics \cite{Fujii20} in basketball). 
In such a case, it may be possible to collect similar plays in the form of a recommendation system based on unsupervised learning (in Section \ref{ssec:unsupervised}), such as in an analogous way of a search on a web page.

Regarding the short-term future prediction discussed in Section \ref{sec:simulate}, these can visually present e.g., how will a certain move work for a player in the same situation as a good player, and how will the team in the next game respond to our team. 
In long-term prediction, predicting the game situations and results of the opponent team in the next game would be useful for tactical planning purposes.
Although there are gaps between the resolution of practical application and research on tactical planning in invasive sports (e.g., formations and styles in soccer \cite{Beal20} and specific cooperative plays and defense styles in basketball \cite{Fujii20}), other team sports such as baseball have fewer such gaps \cite{Stephan18} because most of their actions can be evaluated discretely.  
Since individual results can be more easily predicted in invasive team sports (especially those near the ball), many studies in recent years have evaluated the behaviors of professional athletes (e.g., \cite{Decroos19,Pappalardo19}).

Three possible future issues can be considered. 
One is the high cost of using location information, which currently limits its usage to professional sports. 
This problem is being researched with respect to both software and hardware, and we expect that it may become easier to obtain and more accurate in the future, even for estimating joint positions \cite{Felsen17}. 
With greater spatiotemporal resolution, skillful maneuver in terms of their cognition, force, and torque can be analyzed as described in Section \ref{ssec:handcrafted}.
The second is that higher (almost perfect) performance is often required for practical use. However, it may be more constructive to consider whether the results obtained by these approaches are better (less expensive with fewer mistakes) than those obtained by humans.

\section{Conclusions}
This survey focused on data-driven analyses that can be used to obtain a quantitative understanding of invasion team sports behaviors. 
Two approaches for understanding these multi-agent behaviors were introduced: (1) the extraction of features or rules from data in interpretable ways and (2) the generation and control of behaviors in visually-understandable ways.
Lastly, the potential practical applications of extracted rules, features, and generated behaviors were discussed.  
The development of these approaches would contribute to a better understanding of multi-agent behaviors in the real world.

\acknowledgements
I would like to thank Atom Scott, Masaki Onishi, and Rory Bunker for their valuable comments on this work.
This work was supported by JSPS KAKENHI (Grant Numbers 19H04941, 20H04075, 20H04087) and JST Presto (Grant Number JPMJPR20CA).



\end{document}